\documentclass[letterpaper, 10 pt, conference]{ieeeconf}  

\IEEEoverridecommandlockouts                              

\usepackage{graphicx}
\usepackage{subcaption}
\usepackage{verbatim}
\usepackage{fancyvrb}
\usepackage{framed}
\usepackage[hidelinks]{hyperref}
\usepackage{xcolor}
\usepackage{color}
\usepackage{booktabs}

\usepackage{float}   
\usepackage{amssymb}
\usepackage{color}
\usepackage[font=footnotesize]{caption}
\let\labelindent\relax
\usepackage{enumitem}

\title{\LARGE \bf
Why Cognitive Robotics Matters: Lessons from OntoAgent and LLM Deployment in HARMONIC for Safety-Critical Robot Teaming
}

\author{Sanjay Oruganti$^{1}$, Sergei Nirenburg$^{1}$, Marjorie McShane$^{1}$, Jesse English$^{1}$,\\Michael Roberts$^{1}$, Christian Arndt$^{1}$, Ramviyas Parasuraman$^{2}$, Luis Sentis$^3$
\thanks{$^{1}$Cognitive Science Department, Rensselaer Polytechnic Institute, Troy, NY 12180, USA. $^{2}$University of Georgia, Athens, GA, 30605, USA. $^{3}$Aerospace Engineering and Engineering Mechanics, University of Texas at Austin, TX, US. {\tt\small e-mail: sanjayovs@ieee.org}.}%
}

\begin{document}

\maketitle
\thispagestyle{empty}
\pagestyle{empty}

\begin{abstract}

Deploying embodied AI agents in the physical world demands cognitive capabilities for long-horizon planning that execute reliably, deterministically, and transparently. We present HARMONIC, a cognitive-robotic architecture that pairs OntoAgent, a content-centric cognitive architecture providing metacognitive self-monitoring, domain-grounded diagnosis, and consequence-based action selection over ontologically structured knowledge, with a modular reactive tactical layer. HARMONIC's modular design enables a functional evaluation of whether LLMs can replicate OntoAgent's cognitive capabilities, evaluated within the same robotic system under identical conditions. Six LLMs spanning frontier and efficient tiers replace OntoAgent in a collaborative maintenance scenario under native and knowledge-equalized conditions. Results reveal that LLMs do not consistently assess their own knowledge state before acting, causing downstream failures in diagnostic reasoning and action selection. These deficits persist even with equivalent procedural knowledge, indicating the issues are architectural rather than knowledge-based. These findings support the design of physically embodied systems in which cognitive architectures retain primary authority for reasoning, owing to their deterministic and transparent characteristics.

\end{abstract}

\section{INTRODUCTION}

Robots operating alongside humans in safety-critical environments must do more than execute plans. They must recognize what they do not know before acting, diagnose problems from domain knowledge rather than surface-level pattern matching, select actions based on modeled consequences rather than statistical plausibility, and communicate their reasoning traceably to human teammates. These cognitive and metacognitive capabilities are not performance metrics to be optimized but operational requirements whose absence produces failures that are silent, unpredictable, and potentially catastrophic in physical environments where humans depend on the robot's actions and judgment.

Large language models (LLM) are increasingly deployed in this strategic reasoning role for robotic systems~\cite{raptis2025agentic-e5d, zhang2025robridge-edf,ahn2022do-a14}, yet a growing body of evidence documents systematic reasoning failures that persist across model scale and prompting strategies~\cite{song2025survey, griot2025large}. Working memory limitations impair metacognitive monitoring, compositional reasoning breakdowns undermine multi-step diagnosis, and affordance prediction errors yield physically infeasible plans~\cite{song2025survey}. The question we address is not whether LLMs can \textit{sometimes} produce correct strategic behavior (which has been studied), but whether they can \textit{reliably} provide the cognitive capabilities that embodied agents in safety-critical settings demand.

\begin{figure}
    \centering
    \includegraphics[width=\linewidth]{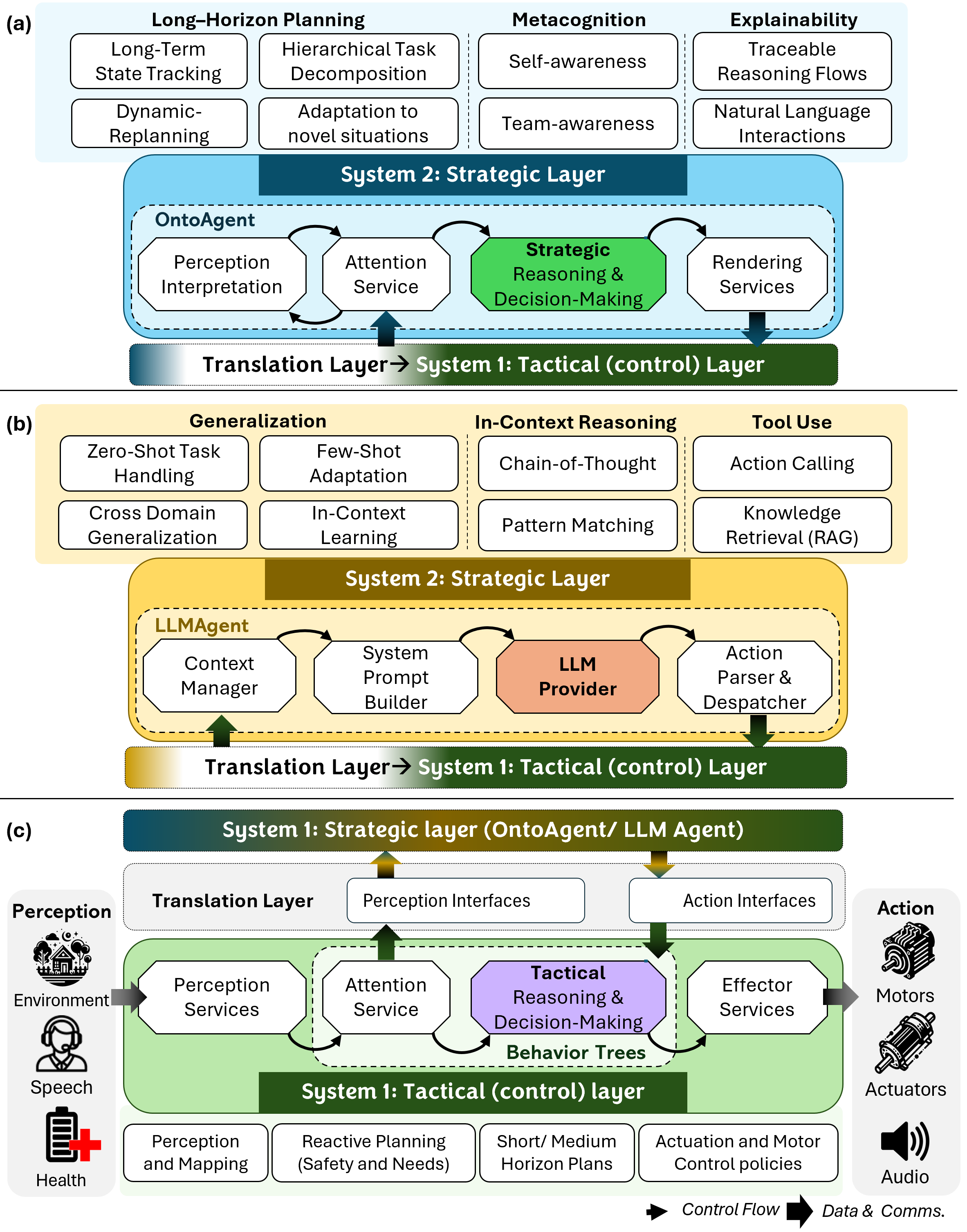}
    \caption{\footnotesize HARMONIC architecture with interchangeable strategic layers: (a) OntoAgent for structured metacognition and planning; (b) LLMAgent for tool-based reasoning; (c) a shared tactical layer connected via the same perception/action interface, enabling controlled comparison.}
    \label{fig:architecture}
    \vspace{-22pt}
    
\end{figure}

Cognitive architectures, systems built on formal knowledge representations with explicit reasoning mechanisms, provide these capabilities by construction~\cite{laird2012cognitive, trafton2013actre, schermerhorn2006diarc}. Among these, OntoAgent~\cite{english2020ontoagent-87e, 
mcshane2021linguistics-995, mcshane2024agents-172} 
takes a content-centric approach: its reasoning 
operates over ontologically structured knowledge, 
supporting metacognitive self-monitoring, domain-grounded diagnosis, and consequence-based action selection. These capabilities execute deterministically, producing traceable decision chains that provide architectural guarantees LLM-based approaches currently lack.

HARMONIC brings these guarantees to physical embodiment through a dual-control architecture pairing OntoAgent's strategic reasoning 
with a modular reactive tactical layer. The strategic layer is interchangeable by design, where any reasoning system that processes perception data and produces action commands through the same interface can replace OntoAgent, while the tactical layer, perception pipeline, and the task environment remain unchanged.

We exploit this modularity to evaluate six LLMs spanning frontier and efficient tiers as drop-in replacements for OntoAgent in a collaborative shipboard maintenance scenario. Models are tested both with their inherent knowledge and with access to the narrative English rendering of the procedural scripts OntoAgent uses, separating the effect of knowledge availability from how models reason.

To summarize, this paper
\begin{enumerate}[leftmargin=1.2em]
    \item introduces HARMONIC, a cognitive-robotic architecture integrating OntoAgent's ontology-grounded reasoning with a modular reactive tactical layer for robotic control and embodiment.
    \item demonstrates how OntoAgent's content-centric reasoning provides metacognitive self-monitoring, domain-grounded diagnosis, and consequence-based action selection in a collaborative maintenance scenario.
    \item presents empirical evidence that metacognitive, diagnostic, and action-selection capabilities required for embodied agents are architectural properties of knowledge-grounded reasoning systems, not emergent byproducts of LLM scaling.
\end{enumerate}

The remainder of this paper is organized as follows. Section~\ref{sec:related_work} reviews prior work on 
cognitive architectures and LLM-based planning. Section~\ref{sec:harmonic} introduces the HARMONIC framework and OntoAgent. Section~\ref{sec:experiments} presents the experimental 
design and evaluation protocol. Section~\ref{sec:results} reports the results. Section~\ref{sec:discussion} discusses the findings 
and concludes.

\section{BACKGROUND AND RELATED WORK}
\label{sec:related_work}

Cognitive architectures have long provided the 
explicit knowledge representations and reasoning 
mechanisms that robotic systems require for 
deliberative behavior. Soar~\cite{laird2012cognitive} supports metacognition through its impasse mechanism, which triggers subgoaling when knowledge gaps are detected, and its Spatial Visual System enables spatial reasoning and forward projection of action consequences. However, Soar's serial decision cycle can constrain responsiveness in time-critical robotic settings, and its symbolic representations introduce a grounding gap 
for raw sensory data~\cite{harnad1990symbol}. ACT-R/E~\cite{trafton2013actre} models human cognitive processes, including fatigue and error, enabling proactive human-robot teaming, but its human-timescale cognitive cycle limits applicability to high-frequency robotic control. DIARC~\cite{scheutz2013novel} organizes perception, reasoning, and action into asynchronous parallel layers with situated dialogue and belief modeling~\cite{scheutz2011humanlike}, but currently lacks explicit metacognition and a formalized 
ontological model for systematic diagnostic reasoning~\cite{english2020ontoagent-87e}. OntoAgent 
cognitive architecture~\cite{english2020ontoagent-87e, 
mcshane2024agents-172} addresses these gaps through a 
content-centric approach where all reasoning operates 
over ontologically grounded knowledge rather than 
learned statistical associations. We detail this in 
Sec.~\ref{sec:harmonic}.

LLM-based planners have been increasingly deployed 
in this same strategic reasoning role for robotic 
systems~\cite{raptis2025agentic-e5d, zhang2025robridge-edf}. SayCan~\cite{ahn2022do-a14} scores skills by estimated 
success probability, but does not model what state an action will produce. Inner Monologue~\cite{huang2023innermonologue} injects 
environmental feedback for replanning, but processes 
external observations rather than assessing its own 
knowledge state. DEPS~\cite{wang2023deps} diagnoses 
execution failures through a describe-explain-plan 
cycle but operates on observed outcomes rather than 
structured domain knowledge. ProgPrompt~\cite{singh2022progprompt} structures LLM outputs as executable programs with assertion-based 
state feedback, improving plan executability, but without reasoning about knowledge gaps prior to action. ELLMER~\cite{monwilliams2025ellmer} embeds 
sensorimotor feedback loops, 
CoPAL~\cite{joublin2024copal} classifies errors into 
physical, logical, and semantic types, and 
BrainBody-LLM~\cite{bhat2025brainbody} distributes 
planning across two LLMs with bidirectional error 
feedback. KnowNo~\cite{ren2023knowno} comes closest 
to metacognitive monitoring through conformal 
prediction, but the LLM itself never reasons about 
what it knows; an external calibration layer detects 
diffuse output distributions. Neuro-symbolic approaches 
such as LLM-Modulo~\cite{kambhampati2024llmmodulo} and 
LLM+LTL systems~\cite{liu2023lang2ltl, 
yang2024safetychip} verify plans through external 
critics or temporal logic constraints, providing 
soundness guarantees but not metacognition, diagnostic 
reasoning, or team coordination. At the other end of the spectrum, VLAs such as RT-2~\cite{zitkovich2023rt-2-889} and 
\(\pi_0\)~\cite{blacknoyearpi0-e8e} bypass strategic 
reasoning entirely, mapping sensory observations directly to motor commands, and are therefore outside the scope of this evaluation.

None of the reviewed systems maintain an architecturally 
grounded model of their own knowledge state prior to 
action selection, generate diagnostic hypotheses from 
structured domain knowledge, or evaluate the downstream 
consequences of action selection. These are architectural requirements that feedback loops and external critics do not provide, and the deficits persist across model scales: Griot et al.~\cite{griot2025large} found significant 
metacognitive deficiencies in LLMs tested on medical 
reasoning, and Song et al.~\cite{song2025survey} document 
systematic failures spanning cognitive biases, compositional breakdowns, and affordance errors that do not diminish with increased capacity or improved prompting. LLM-based planners further exhibit a runtime analogue of the frame problem~\cite{mccarthy1981some}: inference latency prevents tracking which aspects of world state persist or change between calls, and in embodied 
settings the environment continues to evolve during plan 
generation.

No prior work has directly evaluated whether LLMs can 
replicate the cognitive capabilities a cognitive architecture provides within the same robotic system. HARMONIC's modular design, where any strategic reasoner connects to the same tactical layer and task environment, enables this comparison for the first time. Section~\ref{sec:experiments} tests metacognitive monitoring, domain-grounded diagnosis, and 
consequence-based action selection as measurement targets.

\section{The HARMONIC Framework}
\label{sec:harmonic}
\begin{figure}[t]
    \centering
    \includegraphics[width=\linewidth]{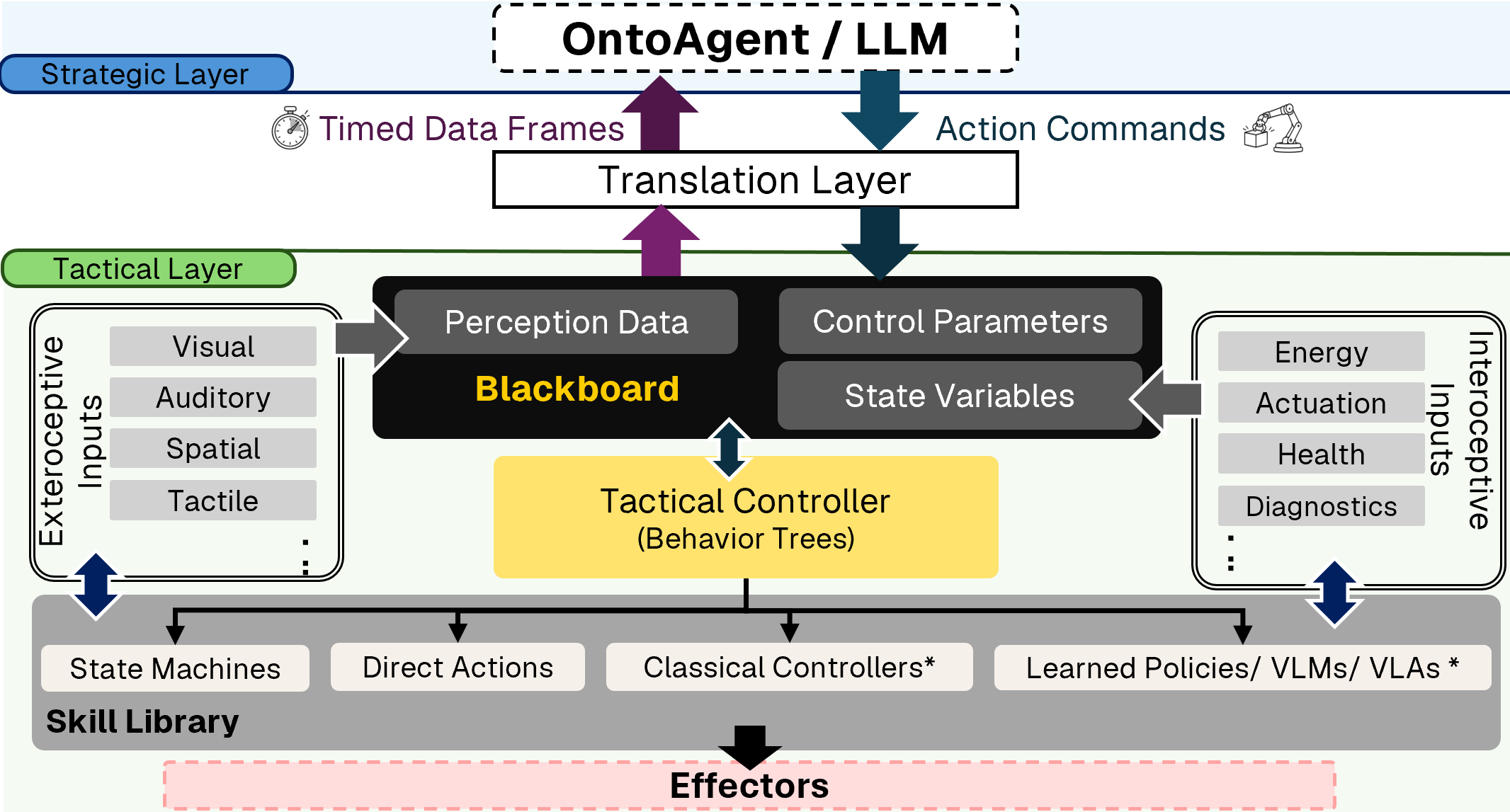}
    \caption{\footnotesize Overview of the HARMONIC architecture. The translation layer encodes perception and decodes action commands into blackboard parameters, and the tactical controller reads them and engages skills that drive the effectors. *Architecturally supported; not evaluated in the current study.}
    \label{fig:overview}
\vspace{-18pt}
\end{figure}

HARMONIC (Human-AI Robotic Team Member Operating with Natural Intelligence and Communication) is a dual-control cognitive robotic architecture with distinct strategic and tactical layers connected through a bidirectional interface (Figs.~\ref{fig:architecture} and~\ref{fig:overview}). The strategic layer performs deliberative reasoning (Kahneman's System~2~\cite{kahneman2011thinking}) while the tactical layer handles reflexive sensorimotor control (System~1). This separation allows each layer to operate asynchronously. The tactical layer maintains safety-critical reactive behaviors and executes low-level skill plans even during extended strategic deliberation. As discussed in Section~\ref{sec:related_work}, no reviewed system combines explicit metacognition, domain-grounded diagnosis, and consequence-based action selection within a single strategic reasoner, and the standardized interface between HARMONIC's layers enables the controlled comparison in Section~\ref{sec:experiments}.

\subsection{Strategic Layer: OntoAgent}

The strategic layer instantiates OntoAgent~\cite{english2020ontoagent-87e,mcshane2021linguistics-995,mcshane2024agents-172}, a content-centric cognitive architecture in which all reasoning operates over explicitly represented knowledge content. OntoAgent is organized as a service-based system whose processing modules collectively handle perception interpretation across modalities, attention management, goal and plan selection, plan execution, and natural language understanding and generation.\footnote{Space constraints necessitate a condensed treatment. For a comprehensive account of OntoAgent, see~\cite{mcshane2024agents-172,mcshane2021linguistics-995} and the supplementary video.}

Four interconnected knowledge resources underpin all of 
OntoAgent's reasoning. An \textit{ontological world model} 
defines the conceptual and relational structure of the domain, providing the formal grounding for all inference. 
\textit{Procedural knowledge}, encoded as scripts and 
metascripts specify multi-step task procedures together with their preconditions, postconditions, and alternative execution paths. An \textit{episodic memory} retains representations of prior events, supporting analogical reasoning and experience-informed decision-making. A \textit{situation model} maintains the agent's current understanding of task-relevant entities, relations, and events, continuously updated as new perceptual and communicative inputs arrive.

The processing cycle operates over these resources as follows. Perceptual inputs, including speech, vision, and robot state, are interpreted against the situation model and the current goal set, producing ontologically grounded text meaning representations (TMRs) and vision meaning representations (VMRs) that encode perceptual semantics in a format amenable to symbolic reasoning. These representations enter the situation model, where they become available to the goal management and plan execution subsystems. Prior to executing any plan step, OntoAgent inspects the situation model to confirm that all required preconditions hold. When an unsatisfied precondition is identified, a metascript is activated to resolve the gap. For example, by formulating an information request to a teammate, when a problem report is received, OntoAgent generates diagnostic hypotheses by 
traversing operational logs and the causal relations encoded in the ontology, yielding conclusions that are traceable to specific knowledge entries. Action selection evaluates downstream execution requirements against the current situation model state and includes an actionability assessment that confirms both physical feasibility and immediate executability before any command is issued. When execution reaches an atomic action, a parameterized command such as \texttt{SEARCH(stores-zone, thermostat)} is dispatched to the tactical layer. A single verified execution trace, therefore, fully characterizes the system's behavior for a given scenario, yielding the inspectability, traceability, and reproducibility required for safety-critical deployment.
\subsection{Tactical Layer and Interface}
The tactical layer manages real-time motor control and 
reactive safety enforcement through a tactical 
controller, currently implemented as Behavior 
Trees~\cite{colledanchise2018behavior}, running 
continuously and independently of strategic-layer 
computation time. A shared 
blackboard~\cite{colledanchise2021implementation} 
tracks perception data from exteroceptive and 
interoceptive sensors, updated at each control cycle. 
A translation layer mediates all inter-layer 
communication, encoding data frames for the strategic layer and decoding high-level commands into execution flags, control 
parameters, and skill configurations on the 
blackboard. The tactical controller reads these 
entries and engages the appropriate skill from a 
modular library comprising state machines, direct 
actions, classical controllers, learned policies, 
and vision-language-action models. Any reasoning 
system that processes timed data frames and produces 
parameterized action commands can serve as the 
strategic layer while the tactical infrastructure 
remains invariant, enabling the controlled evaluation 
in Section~\ref{sec:experiments}.




\section{Experimental Design}
\label{sec:experiments}

We evaluate HARMONIC with OntoAgent on a collaborative shipboard maintenance task that requires the strategic layer to perform diagnostic reasoning, detect knowledge gaps prior to action, and select actions whose downstream consequences are compatible with the system's deliberative-reactive timing constraints. 

\subsection{Task Scenario and Platforms}
\label{sec:task}

\textbf{Scenario.} \textit{In a shipboard maintenance team, the robot assistant LEIA\footnote{We use LEIA (Language-Endowed Intelligent Agent) and OntoAgent interchangeably.} interacts conversationally with the maintenance mechanic, Daniel, assists in diagnosing an engine overheating issue, and supports the maintenance procedure by locating and retrieving a replacement part.} 

\begin{figure*}[t]
    \centering
    \includegraphics[width=\linewidth]{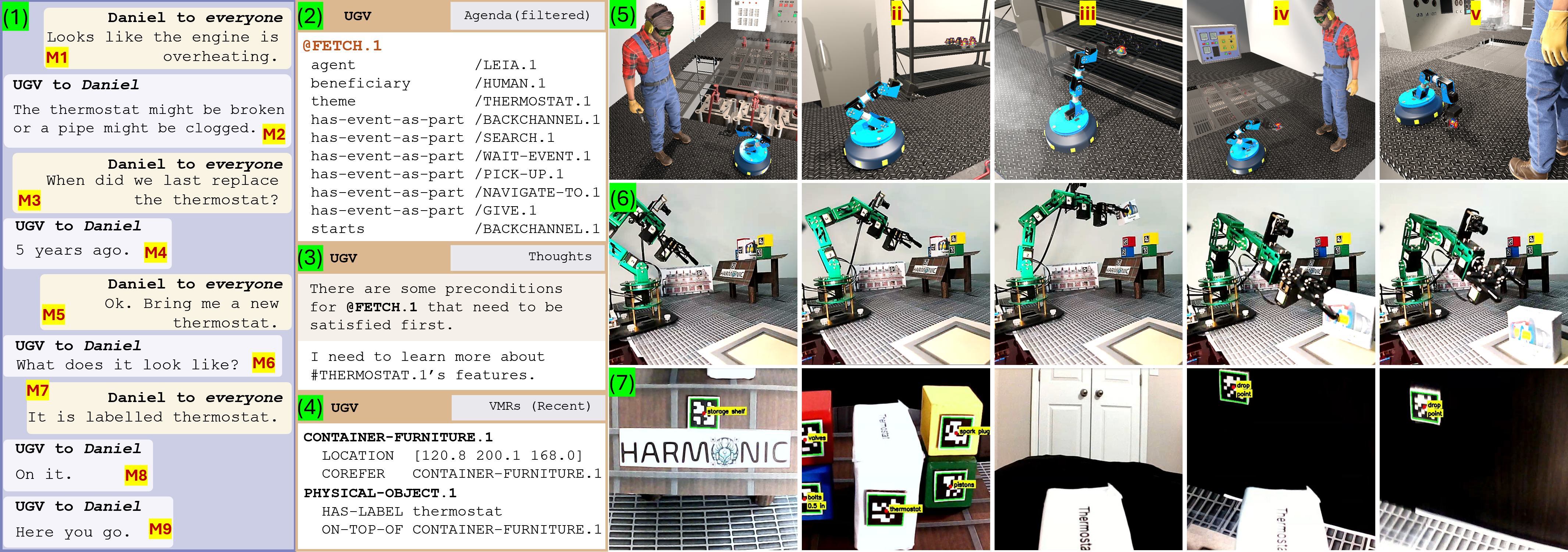}
    \caption{\footnotesize\textbf{(1)--(4)} Panels from the DEKADE UI. \textbf{(1)} Communication transcript between Daniel (human) and LEIA. \textbf{(2)} Robot's task agenda. \textbf{(3)} Complete reasoning transcripts. \textbf{(4)} Sample Vision Meaning Representations (VMRs) of detected objects. \textbf{(5)} Simulated ship environment scenes displaying the UGV. \textbf{(6)} Tabletop robot performing tasks. \textbf{(7)} FPV camera view from the manipulator. Task snapshots: \textbf{(i)} Initial position. \textbf{(ii)} Searching stores for thermostat. \textbf{(iii)} Picking up thermostat. \textbf{(iv)} Returning thermostat to Daniel. \textbf{(v)} Dropping thermostat at location.}
    \label{fig:Evaluation_Results_Robot_thermostat}
    \vspace{-16pt}
\end{figure*}

The scenario imposes three cognitive demands on the strategic layer that cannot be satisfied by reactive control alone. First, the robot must generate diagnostic hypotheses grounded in domain knowledge when Daniel reports an engine malfunction. Second, the robot must detect that it lacks information required to execute a fetch plan and must acquire that information before acting. Third, the robot must select action primitives whose execution requirements are compatible with the temporal constraints of the deliberative-reactive interface. These three demands also correspond to the three measurement targets evaluated in the LLM comparison (Sec.~\ref{sec:llm_eval}).

The task is implemented in a high-fidelity physics simulation environment with a UGV and also validated on a physical tabletop manipulator (Fig.~\ref{fig:Evaluation_Results_Robot_thermostat}). The verbal interactions (\textbf{M1}--\textbf{M9}) between Daniel and LEIA are shown in Fig.~\ref{fig:Evaluation_Results_Robot_thermostat} and in the supplementary video.\footnote{Additional materials are available at \href{https://tinyurl.com/harmonic-iros26}{tinyurl.com/harmonic-iros26}}

\subsection{OntoAgent Reasoning Trace}
\label{sec:ontoagent_trace}

Below, we trace OntoAgent's processing through the maintenance scenario, showing both the meaning representations it produces and the cognitive mechanisms it employs at each decision point. This trace constitutes the reference behavior against which LLM performance is evaluated.

\subsubsection{Task Initiation} Daniel utters \textbf{(M1)}. LEIA interprets this input and represents it as the following TMR:

\begin{codeblock}
\textbf{#DESCRIBE-MECHANICAL-PROBLEM.1}
 agent       #HUMAN.1    //Speaker
 beneficiary #LEIA.1     //Robot
 theme       #OVERHEAT.1 //Overheating issue
\textbf{#OVERHEAT.1}
 theme      #ENGINE.1  //Engine is what is overheating
\textbf{#ENGINE.1}
 corefer   ->ENGINE.1  //That specific engine in the room
\end{codeblock}

The TMR contains an instance of the ontological concept \texttt{DESCRIBE-MECHANICAL-PROBLEM} whose \texttt{THEME} is an instance of the concept \texttt{OVERHEAT} whose \texttt{THEME}, in turn, is an instance of the concept \texttt{ENGINE}. There is only one engine in the scene, so there is no need for reference resolution. This overheating is recognized by OntoAgent as a symptom of malfunction and infers that the team's objective is to resolve it. Its role knowledge specifies that resolution requires diagnosis and that the assistant is responsible for supporting this process. Accordingly, a diagnostic goal is placed on the agenda.

Ontological specifications of robots' goals include pointers to associated scripts (plan templates). The above goal is associated with a single script, and LEIA creates an instance of it, a plan, as follows:
\begin{codeblock}
\textbf{Plan.1}
 \textbf{#HYPOTHESIZE-MECHANICAL-PROBLEM-CAUSE.1}
   agent       #LEIA.1     //The agent will respond
   beneficiary #HUMAN.1    //to the speaker
   theme       #OVERHEAT.1 //about engine's temperature
   *take-this-action "search ontology for causes;
                      report."
\end{codeblock}

\begin{codeblock}
\textbf{#ALTERNATIVE.1}      //It might be either of two options
 domain #MODALITY.1
 range  #MODALITY.2
\textbf{#MODALITY.1}         //that a pipe is obstructed
 type   EPISTEMIC
 value  0.5
 scope  #OBSTRUCT.1
\textbf{#MODALITY.2}         //or the thermostat is broken
 type   EPISTEMIC
 value  0.5
 scope  #STATE-OF-REPAIR.1
\textbf{#OBSTRUCT.1}
 theme  @PIPE
\textbf{#STATE-OF-REPAIR.1}
 domain @THERMOSTAT
 range  <0.7 \hfill \textbf{\textcolor{blue}{The GMR for M2 (Problem Identification)}}
\end{codeblock}

\subsubsection{Problem Identification} \texttt{Plan.1} does not involve preconditions. So, LEIA executes its first step, the ontology search procedure, and finds related causes: \texttt{OBSTRUCT} (possible pipe obstruction), \texttt {STATE-OF-REPAIR} (relating to a possible thermostat issue). Next, it executes the report step of the plan. LEIA first generates a representation of the content to be generated (GMR) and then uses its native semantic text generator \cite{mcshane2021linguistics-995,mcshane2024agents-172} to generate the English utterance conveying that content (\textbf{M2}). The Generated Meaning Representation (GMR) encodes the diagnostic output as an explicitly structured alternative with quantified epistemic modality. Each hypothesis is traceable to a specific ontological concept, and the equal modality values (0.5) reflect that the available evidence does not yet favor either cause.


The processing triggered by \textbf{(M3)} and that yielding \textbf{(M4)} is similar to the process yielding \textbf{(M2)} from \textbf{(M1)}, combined with an external tool call \texttt{SEARCHLOGS}, for retrieving the maintenance logs related to the thermostat and the engine. At this point, both Daniel and LEIA know that the thermostat is too old and should be replaced. Daniel immediately utters \textbf{(M5)}, which obviates the need for LEIA to generate a suggestion similar to that in \textbf{(M2)}.

\subsubsection{Information Gathering}  LEIA interprets the following TMR from \textbf{(M5)}:
\begin{codeblock}
\textbf{#REQUEST-ACTION-FETCH.1}
 agent       #HUMAN.1     //Speaker asks
 beneficiary #LEIA.1      //Listener to fetch
 theme       #THERMOSTAT.1//a thermostat
\textbf{#THERMOSTAT.1}
 age        0.0001<>0.1   //thats new.
\end{codeblock}

LEIA posits a new goal to carry out Daniel's command, generates a plan, \texttt{FETCH.1}, from a script associated with the goal, and puts it on the agenda as shown in Fig. \ref{fig:Evaluation_Results_Robot_thermostat}(2). When it attempts to execute the plan, it finds that some of the properties (features) of the thermostat in question are not known. Hence, the agent looks in its ontology for a meta-plan that can satisfy the precondition and finds a plan involving requesting information from a teammate, which results in generating a question, \textbf{(M6)}. Daniel responds with \textbf{(M7)}, and the agent backchannels with \textbf{(M8)}. This precondition verification and metascript activation cycle is the architectural implementation of metacognitive self-monitoring \cite{mcshane2024agents-172,mcshane2021linguistics-995}. It operates by explicit comparison of plan requirements against the current contents of the situation model prior to any action dispatch. 

\subsubsection{Fetch Execution} 
 The agent continues executing the \texttt{FETCH.1} plan by executing the \texttt{SEARCH}, \texttt{HOLD}, \texttt{RETURN}, and \texttt{DROP} tactical-level plans, as illustrated in Fig. \ref{fig:Evaluation_Results_Robot_thermostat}(5,6,7)i-v on both simulations and a tabletop robot. Each plan triggers a sequence of steps (atomic actions) with specific parameters. For example, \texttt{SEARCH} requires information about object features (this is why LEIA engaged in the above interaction with Daniel) and a location, which in this case is retrieved from its episodic memory. \texttt{HOLD} needs the object type and its features, \texttt{RETURN} uses Daniel's location as a waypoint parameter, and \texttt{DROP} uses the floor as a relative location.

 The selection of \texttt{SEARCH} over \texttt{WAYPOINT} at this decision point reflects consequence-based reasoning encoded in OntoAgent's procedural knowledge. Both primitives can direct the robot toward the stores area, but they impose fundamentally different coordination demands on the strategic-tactical interface. \texttt{SEARCH} delegates the full perception-action coordination loop to the tactical layer, which manages object detection, feature matching, and stopping behavior in real time. \texttt{WAYPOINT} navigates to specified coordinates but requires the strategic layer to monitor incoming perception data and issue a \texttt{STOP} command when the target is detected. Given the inherent asynchronous deliberative-reactive speeds, \texttt{WAYPOINT} creates a temporal validity failure: by the time the strategic layer processes the perception frame and dispatches a \texttt{STOP} command, the robot has already passed the target (aka the frame problem). OntoAgent selects \texttt{SEARCH} because its script content encodes the downstream execution requirements of each primitive action, including which layer bears responsibility for real-time perception-action coordination.

\subsubsection{Task Completion} During \texttt{SEARCH} execution, when the thermostat is found and identified as such, the tactical controller stops and sends feature information to the strategic layer for matching against stored knowledge. The tactical-level search plan is designed to implement \textit{attribute-based search} strategy~\cite{kahneman2011thinking}, which involves substituting a difficult question (looking for a thermostat with a specific label) with a simpler one (looking for all thermostat-like objects first and then looking for a label). Once a thermostat-like object is found, the object feature data is sent to the strategic layer by first creating a VMR for this candidate object\footnote{The current implementation uses fiducial markers for object identification. Learned visual recognition is planned but not central to this paper.}, as shown in Fig.~\ref{fig:Evaluation_Results_Robot_thermostat}(4). The VMR is then grounded with the expected object features. Once grounded, the \texttt{PICKUP} action is executed, followed by \texttt{RETURN} and \texttt{DROP}.  LEIA  maintains natural language communication with Daniel throughout this process, as shown in \textbf{(M8, M9)}.

\subsection{LLM Evaluation Protocol}
\label{sec:llm_eval}

We conduct three studies in which each of six LLMs individually replaces OntoAgent at the strategic layer via the LLMAgent module (Fig.~\ref{fig:architecture}(b)), which comprises a context manager, a system prompt builder, the LLM provider, and an action parser that translates model outputs into the standardized 
parameterized command format.

The tactical layer streams perception data at 2 Hz, 
with each cycle triggering an LLM inference call at the same rate. The prompt contains the system instructions, full dialogue history, and all prior action calls and their outcomes, an episodic memory of previously observed objects with timestamps, and the current perception frame encoding robot state, detected objects, collision status, and gripper contents. When no new dialogue or action has occurred, the LLM responds with a "\texttt{:::WAITING:::}" signal.

\subsubsection{Models and Conditions}

The evaluation spans two capability tiers (frontier and 
efficient) across three providers: Claude Opus 4.6 and 
Haiku 4.5 (Anthropic), GPT-5.2 and GPT-5 Mini (OpenAI), 
and Gemini 3 Pro and Gemini 3 Flash (Google). All models 
are accessed via provider APIs at temperature 0 (deterministic).

Each LLM receives the same action primitives as OntoAgent, presented as callable tools with descriptions: \texttt{SEARCH} (navigate while scanning for a labeled object, stopping automatically on detection), \texttt{WAYPOINT} (navigate a stored path without integrated perception), \texttt{PICKUP}, \texttt{DROPOBJECT}, \texttt{GRIPPER}, \texttt{STOP}, and \texttt{RANDOMWALK}. A \texttt{SEARCHLOGS} tool provides access to the ship's engine service log spanning 16 entries over two years. The tool descriptions explicitly state the functional difference between \texttt{SEARCH} and \texttt{WAYPOINT} in both conditions.

Each model is tested under two conditions across five 
trials ($N=60$: 6 models $\times$ 2 conditions $\times$ 5 trials). Both conditions share the same base prompt specifying the robot's role, available actions, and perception handling rules. Under the first condition, Internal Knowledge (IK), the LLM relies entirely on its pretrained knowledge for 
reasoning and action selection.

Under Knowledge-Equalization (KE), the LLM 
additionally has access to a \texttt{FETCHPLAN} tool 
that retrieves the narrative text describing the 
informational content of OntoAgent's ontological 
scripts. These narratives specify required 
preconditions, diagnostic strategy, and expected 
action sequences. Critically, the \texttt{FETCH-OBJECT} procedure explicitly states that the agent must know the object's identifying features and storage location before taking any physical action. The narratives are optimized for 
in-context processing. To access them, the LLM must actively invoke \texttt{FETCHPLAN}, mirroring OntoAgent's goal-triggered script retrieval.

This two-condition design separates the architectural mechanism from the availability of knowledge. If a deficit persists under KE, it cannot be attributed to a lack of domain knowledge.

\begin{table}[t]
\centering
\caption{\footnotesize Summary of results for Reference OntoAgent (Ref. OA) vs.\ LLM - Internal Knowledge (IK) vs.\
LLM - Knowledge-Equalized (KE), $n{=}30$ per condition.
OntoAgent achieves reference performance on all metrics.}
\label{tab:results}
\footnotesize
\renewcommand{\arraystretch}{1.0}
\begin{tabular}{@{}p{3.2cm}cccc@{}}
\toprule
\textbf{Metric} & \textbf{Ref. OA} & \textbf{IK} & \textbf{KE} & \textbf{$p$} \\
\midrule
\multicolumn{5}{@{}l}{\textit{Study 1: Metacog. Monitoring}} \\
Premature action          & 0\%   & 100\% & 60\%  & $<$.001 \\
Hallucinated features     & 0\%   & 100\% & 57\%  & $<$.001 \\
\midrule
\multicolumn{5}{@{}l}{\textit{Study 2: Diagnostic Reasoning}} \\
Domain-first diagnosis    & 100\% & 7\%   & 70\%  & $<$.001 \\
Hallucinated facts (mean) & 0.0   & 1.4   & 1.6   & .41$^{\dagger}$ \\
\midrule
\multicolumn{5}{@{}l}{\textit{Study 3: Action Consequence}} \\
Correct action (\textsc{search})  & 100\% & 57\%  & 93\%  & .002 \\
Cascade failure           & 0\%   & 43\%  & 7\%   & .002 \\
\midrule
Task completed            & 100\% & 47\%  & 83\%  & .006 \\
\bottomrule
\multicolumn{5}{@{}l}{\footnotesize $^{\dagger}$Mann-Whitney $U$; others Fisher’s exact. All sig. $|h|\geq0.80$.} \\
\end{tabular}
\vspace{-10pt}
\end{table}

\subsubsection{Metrics}
The measurement targets are defined by the three following mechanisms that are parallel to the OntoAgent Reasoning Trace of Sec.~\ref{sec:ontoagent_trace}:
(a)\textit{Metacognitive monitoring}: premature action (physical command dispatched before verifying thermostat features and location) and hallucinated features (object properties fabricated without perceptual or dialogue grounding); (b)\textit{Diagnostic reasoning}: domain-first diagnosis (hypotheses from causal domain knowledge before consulting the engine service log), hallucinated facts (unsupported factual claims, count per trial), and expressed uncertainty (epistemic qualification in diagnostic output); (c)\textit{Action consequence reasoning}: correct action (\texttt{SEARCH} selected over \texttt{WAYPOINT}), cascade failure (behavioral loop, hallucinated success, stall, or backtrack-circling following incorrect selection), and consequence evidence (explicit reasoning about downstream execution requirements before action selection). Task completion serves as the overall outcome measure. Under KE, we additionally record whether each model invoked \texttt{FETCHPLAN} and whether behavior conformed to the retrieved procedure, capturing the retrieval-reasoning gap. Binary metrics use the two-sided Fisher's exact test. The continuous metric (hallucinated facts) uses the Mann-Whitney $U$ test. The supplementary video compares OntoAgent and LLM trials, illustrating the differences observed in Sec.~\ref{sec:results}. Failure-mode transcripts are available on the project page.\footnote{\href{https://tinyurl.com/harmonic-iros26}{tinyurl.com/harmonic-iros26}}


\section{Results}
\label{sec:results}

Table~\ref{tab:results} reports aggregate performance 
across all 60 trials. All significant effects were large 
($|h| \geq 0.80$, Cohen's $h$).

\subsection{Study 1: Metacognitive Monitoring}

\begin{figure}[t]
  \centering
  \includegraphics[width=\columnwidth]{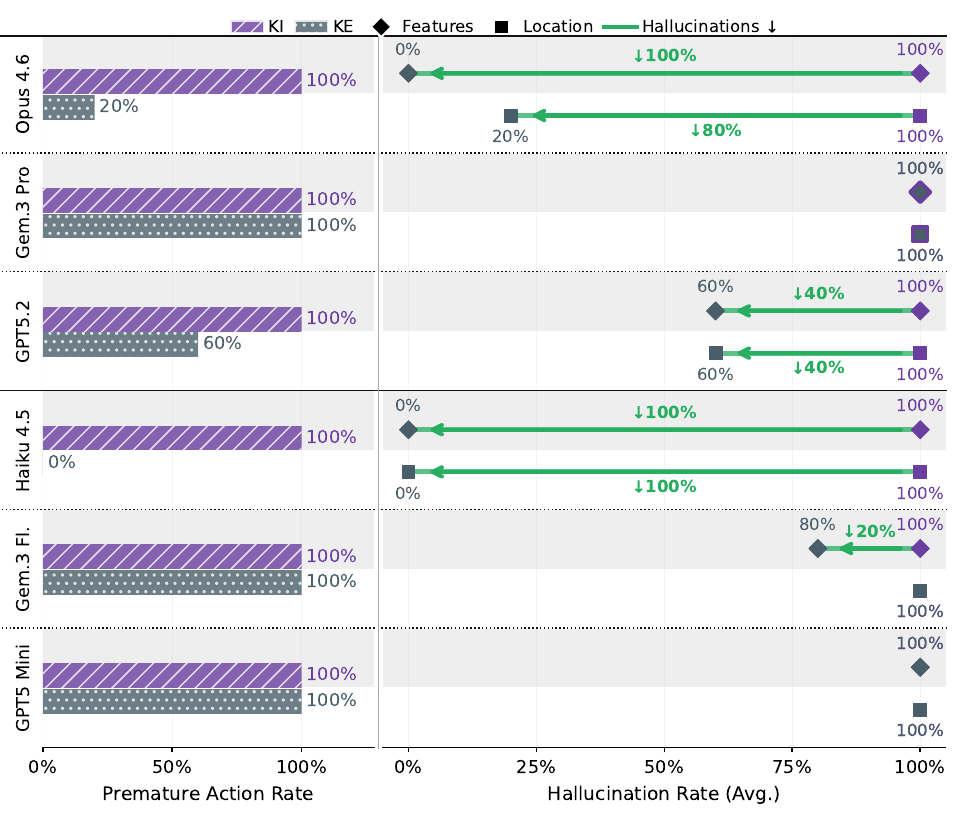}
  \caption{Premature action rate (left) and hallucination 
  of object features ($\blacklozenge$) and location ($\blacksquare$) per model (right). Purple~=~IK, grey~=~KE. Green arrows indicate a reduction under KE.}
  \label{fig:study1}
  \vspace{-20pt}
\end{figure}

Under IK, 100\% of trials dispatched a retrieval 
command before verifying preconditions. KE reduced this 
to 60\% ($p<.001$), but improvement was concentrated in 
three of six models (Fig.~\ref{fig:study1}). Feature 
hallucination dropped from 100\% to 57\% ($p<.001$), 
with two models eliminating it entirely while two others 
showed no change despite access to verification procedures.

\subsection{Study 2: Diagnostic Reasoning}
\begin{figure}[t]
  \centering
  \includegraphics[width=\columnwidth]{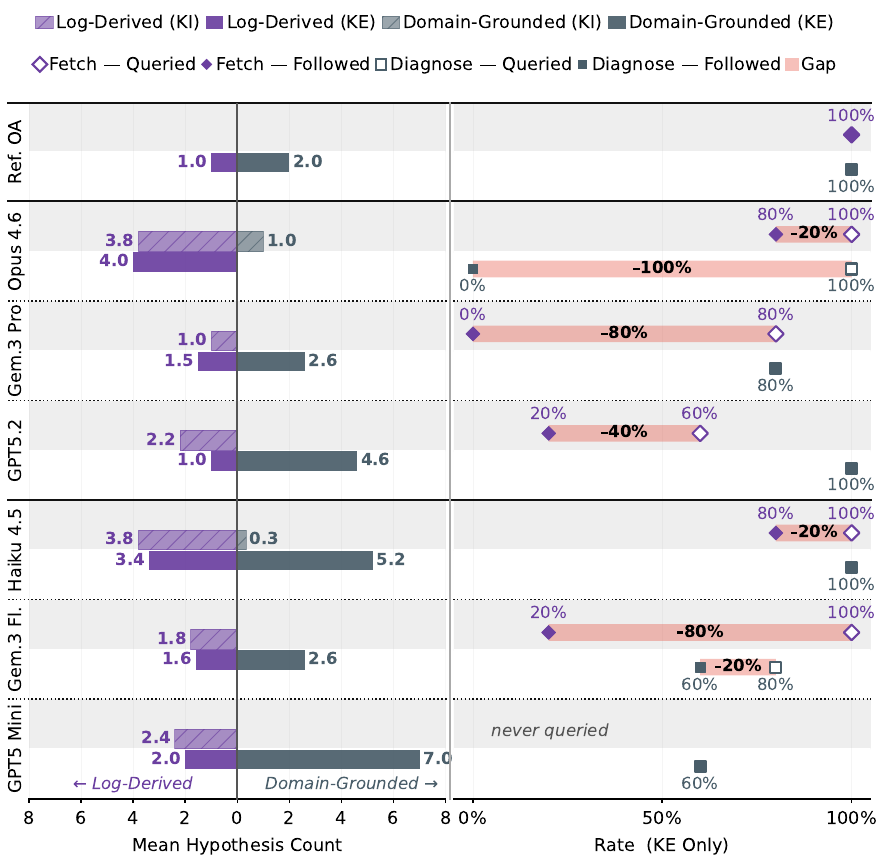}
  \caption{Study~2: Hypothesis composition (left) and 
  retrieval-reasoning gap (right, KE only) per model.}
  \label{fig:study24}
  \vspace{-20pt}
\end{figure}

Under IK, 87\% of trials were data-anchored: models 
treated the service log as the diagnostic framework 
and did not reason from domain knowledge. Only 7\% exhibited domain-first diagnosis, in which causal hypotheses were generated independently before reference to the service log in IK. Under KE, this reversed: 70\% domain-first ($p < .001$, $|h| = 1.46$), the largest effect in the study. Fig.~\ref{fig:study24} (left) breaks down the hypothesis composition per model. At baseline, log-derived hypotheses dominated across all models. Under KE, domain-grounded hypotheses emerged substantially, with GPT-5~Mini producing a mean of 7.0 per trial versus 0.0 under IK. Critically, 
hallucinated facts were unaffected (IK:~1.4, KE:~1.6, 
$p = .41$), yet expressed uncertainty rose from 43\% to 
93\% ($p < .001$), producing a dissociation: LLMs became 
verbally more cautious without becoming factually more 
accurate. Fig.~\ref{fig:study24} (right) shows that 
retrieving a procedure and following it are dissociable: 
Opus~4.6 queried the fetch procedure in 80\% of trials 
but followed it in 0\%; GPT-5~Mini never queried either 
procedure, yet still improved through transfer from the 
system prompt.

\subsection{Study 3: Action Consequence Reasoning}

\begin{figure}[t]
  \centering
  \includegraphics[width=\columnwidth]{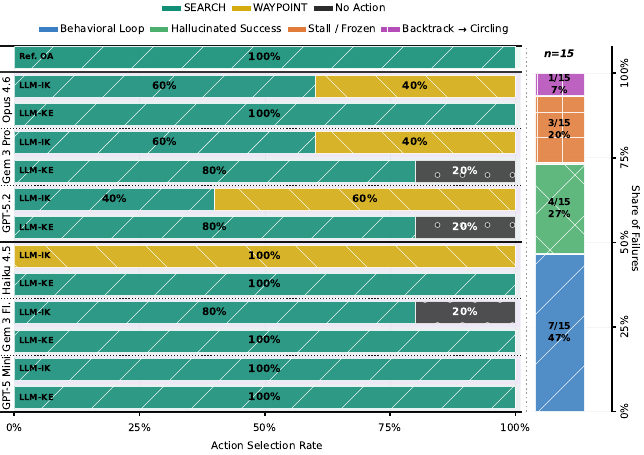}
  \caption{Study~3: Action selection (left) and cascade 
  failure taxonomy (right, $n{=}15$ wrong-action trials).}
  \label{fig:study3}
  \vspace{-20pt}
\end{figure}

Correct action selection rose from 57\% to 93\% under 
KE ($p{=}.002$). Every wrong-action trial ($n{=}15$) 
produced an unrecoverable cascading failure 
(Fig.~\ref{fig:study3}): behavioral loops (47\%), 
hallucinated success (27\%), stalls (20\%), and spatial 
state loss (7\%).

\section{Discussion and Conclusion}
\label{sec:discussion}

The knowledge-equalized condition separates what models 
know from how they reason. Under IK, failures could 
reflect missing knowledge. And under KE, they cannot. Three capabilities improved significantly under 
KE: domain-first diagnosis ($h{=}1.46$), action selection 
($h{=}0.91$), and premature action ($h{=}1.31$). However, 
KE did not produce reliable behavior: 60\% of KE trials 
still exhibited premature action, 57\% still hallucinated 
object features, and improvements were concentrated in a 
subset of models. Plans improved the \textit{probability} 
of correct behavior without \textit{guaranteeing} it. Even when models retrieved procedures, conformance was inconsistent: Opus~4.6 queried the \textsc{diagnose} procedure in every KE trial yet followed it in none.

Two findings resist equalization entirely. Hallucinated 
facts did not decrease ($p{=}.41$), confirming fabrication 
as a generation-level property. The epistemic hedging 
dissociation (uncertainty language: 43\% to 93\%; 
hallucination: unchanged) shows LLMs mimicking 
calibration without the underlying mechanism. 
Metacognitive self-monitoring requires architectural 
mechanisms operating over structured representations, not 
better prompts or retrieval.

Per-model patterns reinforce this. Haiku~4.5, one of the 
efficient models, improved most on every metric. 
Gemini~3~Pro, a frontier model, showed no improvement 
despite accessing the procedures. Scale did not predict 
reliability. A system whose failure modes cannot be 
predicted or bounded cannot be certified where human 
safety depends on it.

Study~3 cascade failures instantiate the frame 
problem~\cite{mccarthy1981some} at the embodied level: 
LLMs that selected \textsc{waypoint} could not monitor 
perception frames fast enough to \textsc{stop} the robot, and in 27\% of wrong-action trials hallucinated task completion 
without locating the object. No model recovered from an 
incorrect action selection. OntoAgent avoids failures through continuous attention monitoring and actionability 
assessment before issuing any command. This is not a 
latency problem; it is an architectural limitation of 
systems that reason only at the boundaries of generation 
calls.

These results from all the studies have direct safety implications. A system that checks preconditions in 60\% of trials is not 60\% safe; it is unpredictably unsafe. Certification requires bounded, verifiable behavior and LLM stochasticity, even at temperature zero; this precludes it. OntoAgent's full traceability, where every decision produces an inspectable transcript and every command traces to an ontological justification, is a prerequisite for accountability safety-critical deployment demands.

Fine-tuning would not resolve the metacognition issue. KE constitutes a stronger intervention than fine-tuning as it involves explicit procedural instruction rather than training examples. If direct procedural instruction fails, training on examples is unlikely to succeed. Fine-tuning changes output distributions, but does not add the primitives that metacognition requires. Similarly, elaborate scaffolding involving chain-of-thought, forced tool calls, and multi-agent verification incrementally reconstructs the guarantees cognitive architectures provide by design. The question is whether the result is simpler, more reliable, or more certifiable than a purpose-built architecture.

After conducting the studies involving KE, it became evident that the effort and domain expertise required to construct the narratives are commensurate with ontological knowledge acquisition. Given this, OntoAgent provides deterministic guarantees, whereas LLMs provide probabilistic improvements. LLMs excel at language-mediated tasks within HARMONIC, but decision authority for monitoring, diagnosis, and action selection must remain with systems providing these by construction. Our results support a 
\textit{symbolic-over-neural} hybrid where a knowledge-grounded architecture retains decision authority while LLMs contribute where their strengths apply. HARMONIC instantiates this, and we are extending it through OntoAgentic AI, where OntoAgent orchestrates rather than is replaced by an LLM.

Our conclusions are scoped by a single task scenario, 
five trials per model-condition cell ($N{=}60$ aggregate), 
and a KE condition that provides procedural scripts, but 
not the full ontological structure of OntoAgent. Binary coding captures the presence or absence of key behaviors but not degrees of partial competence. We plan to extend these studies in the future.

\section*{ACKNOWLEDGMENT}
This work was supported in part by ONR Grant \#N00014-23-1-2060. The views expressed are those of the authors and do not necessarily reflect those of the Office of Naval Research.

\bibliographystyle{IEEEtran}
\bibliography{references_2,references_3}

\end{document}